\documentclass{article}

\usepackage{arxiv}

\usepackage[utf8]{inputenc} 
\usepackage[T1]{fontenc}    
\usepackage{hyperref}       
\usepackage{url}            
\usepackage{booktabs}       
\usepackage{amsfonts}       
\usepackage{nicefrac}       
\usepackage{microtype}      
\usepackage{lipsum}
\usepackage{graphicx}
\usepackage{multirow}
\usepackage{caption}
\usepackage{float}
\usepackage{fancyhdr}
\usepackage{amsmath}
\usepackage{tikz}
\usepackage{geometry}
\usepackage{pgffor}
\graphicspath{ {./images/} }

\title{GRAFT: GRaPH and Table Reasoning for Textual Alignment — A Benchmark for Structured Instruction Following and Visual Reasoning}

\author{
 Abhigya Verma \\
 ServiceNow \\
 \texttt{abhigya.verma@servicenow.com} \\
 \And
 Sriram Puttagunta \\
 ServiceNow \\
 \texttt{sriram.puttagunta@servicenow.com} \\
 \And
 Seganrasan Subramanian \\
 ServiceNow \\
 \texttt{seganrasan.subramanian@servicenow.com} \\
 \And
 Sravan Ramachandran \\
 ServiceNow \\
 \texttt{sravan.ramachandran@servicenow.com} \\
}

\begin{document}
\maketitle
\begin{abstract}
GRAFT is a structured multimodal benchmark designed to probe how well LLMs handle instruction-following, visual reasoning, and tasks requiring tight visual–textual alignment. The dataset is built around programmatically generated charts and synthetically rendered tables, each paired with a carefully constructed, multi-step analytical question that depends solely on what can be inferred from the image itself. Responses are formatted in structured outputs such as JSON or YAML, enabling consistent and fine-grained evaluation of both reasoning processes and adherence to output specifications. The benchmark further introduces a taxonomy of reasoning operations—ranging from comparison and trend identification to ranking, aggregation, proportional estimation, and anomaly detection—to support a comprehensive assessment of model capabilities. Taken together, GRAFT provides a unified and scalable framework for evaluating multimodal LLMs on visually grounded, structured reasoning tasks, offering a more rigorous standard for future benchmarking efforts.
\end{abstract}


\section{Introduction}
\label{sec:intro}

Multimodal foundation models have advanced rapidly in recent years, demonstrating strong performance on a wide spectrum of tasks—from image captioning and visual question answering (VQA) to document understanding and embodied reasoning. Benchmarks such as VQAv2 \cite{goyal2017making}, VizWiz \cite{bigham2010vizwiz}, and DocVQA \cite{mathew2021docvqa} have been instrumental in tracking this progress. Yet, these datasets primarily revolve around natural images or scanned documents and frequently lack the structural complexity or controllable variation needed to probe more specialized reasoning skills. Consequently, they offer only a partial view of how well these models handle fine-grained analytical reasoning over structured visual inputs.

Despite impressive strides in general visual and linguistic competence, contemporary multimodal systems remain insufficiently tested on structured content—particularly charts and tables. Such data demands a blend of perceptual understanding and more advanced analytical abilities, including identifying trends, comparing relationships, and carrying out numerical inferences. Many real-world settings further require models to produce rigorously formatted outputs (e.g., JSON, YAML), adding an additional layer of difficulty as systems must align visual interpretation with structured text generation.

\subsection{Motivation}

To address these gaps, we introduce \textbf{GRAFT}—\textbf{GRaPH and Table Reasoning for Textual Alignment}—a benchmark purpose-built to evaluate multimodal models on structured reasoning tasks involving charts and tables. GRAFT relies on programmatically generated synthetic data: all visual elements are constructed using Python plotting libraries, allowing for precise control over stylistic variation, task difficulty, and dataset composition. While natural data collection can be labor-intensive and susceptible to annotator bias, synthetic generation provides a scalable alternative that mitigates these issues—though it may occasionally introduce minor artifacts inherent to automated pipelines.

This design choice enables a cleaner assessment of reasoning skills by minimizing the noise, irregularities, and uncontrolled variation typically associated with natural datasets.

\subsection{Research Questions}
\label{sec:research_questions}

GRAFT is structured to explore several central questions:
\begin{enumerate}
\item To what degree can current multimodal models carry out structured reasoning when presented with chart- and table-based visual inputs?
\item How well do these models follow complex, multi-step instructions while also producing structured outputs that are both semantically faithful and syntactically valid?
\item Are the characteristic error patterns different for chart reasoning compared to table reasoning?
\item What limitations emerge with respect to visual grounding and the alignment between textual prompts and visual semantics?
\end{enumerate}

By examining these issues in a controlled and transparent setting, GRAFT offers a principled way to diagnose the reasoning capabilities of multimodal systems. In doing so, it highlights key shortcomings and lays the groundwork for future research aimed at more reliable and semantically grounded structured reasoning.

The complete GRAFT dataset is accessible on Hugging Face\footnote{\href{https://huggingface.co/datasets/ServiceNow-AI/GRAFT_benchmark}{https://huggingface.co/datasets/ServiceNow-AI/GRAFT\_benchmark}}
. Appendix \ref{appendix:dataset-preview} provides further details on dataset construction along with illustrative examples.

\section{Related Work}
\label{sec:related_work}
Recent advances in multimodal reasoning span both benchmark development and vision-language model (VLM) architectures. We review (1) the progression of multimodal QA datasets, and (2) instruction-following strategies in VLMs.

\subsection{Multimodal Benchmarks}
Early benchmarks like VQA v2.0 \cite{goyal2017vqa} and COCO-QA \cite{ren2015cocoqa} addressed basic object and attribute queries, but lacked complex reasoning. Later, synthetic datasets such as ChartQA \cite{masry2021chartqa} and PlotQA \cite{plotqa2023} enabled structured reasoning on charts at scale. Document-based datasets like DocVQA and TAT-DQA \cite{tatdqa2024} focused on text extraction, while GRAFT bridges this with added structural complexity and numerical reasoning, similar to TabFact \cite{chen2020tabfact}.

To contextualize GRAFT within the broader landscape of multimodal evaluation, Table \ref{tab:benchmark_comparison} provides a comparative overview of widely used benchmarks and highlights the unique innovations introduced by GRAFT.

\begin{table*}[!htbp]
\centering
\caption{Comparison of Multimodal Benchmarks and GRAFT's Innovations}
\label{tab:benchmark_comparison}
\begin{tabular}{p{2.5cm}p{5cm}p{9cm}}
\toprule
\textbf{Benchmark} & \textbf{Short Description} & \textbf{GRAFT's Improvement} \\
\midrule
CoSyn \cite{yang2024cosyn} & Code-guided synthetic multimodal data for text-rich images & Provides large-scale synthetic instruction-tuning data; GRAFT complements by targeting fine-grained structured reasoning and strict evaluation protocols \\ \hline
VQA v2.0 \cite{goyal2017vqa} & Open-ended QA on natural images & Adds structured reasoning over charts/tables with schema-constrained outputs \\ \hline
GQA\cite{gqa2019} & Scene graph QA for compositional reasoning & Introduces numerical operations (aggregation, proportion) absent in scene graphs \\ \hline
TextVQA\cite{textvqa2020} & OCR-based QA on text-heavy images & Focuses on analytical vs. extraction tasks, with structured answer formats \\ \hline
TallyQA\cite{masry2021tallyqa} & Complex counting with relationships & Extends to multi-step reasoning beyond object counting \\ \hline
ChartQA\cite{masry2021chartqa} & Logical operations on charts & Adds programmatic chart generation with pixel-level control \\ \hline
DocVQA\cite{mathew2021docvqa} & Document understanding from scans & Shifts focus from text extraction to numerical table/chart analysis \\ \hline
ScienceQA\cite{scienceqa2022} & Multimodal science multiple-choice QA & Introduces open-ended structured responses (JSON/YAML) \\ \hline
PlotQA\cite{plotqa2023} & 10M synthetic plot QA pairs & Adds real-world domain grounding and jury-validated QA pairs \\ \hline
MMBench\cite{mmbench2023} & Comprehensive cyclic evaluation & Specializes in fine-grained visual-textual alignment metrics \\ \hline
TAT-DQA\cite{tatdqa2024} & Technical document QA & Combines textual and visual table reasoning in unified framework \\ \hline
M3SciQA\cite{li2024m3sciqa} & Multi-document scientific QA & Focuses on single-document visual reasoning with synthetic data \\ \hline
SceMQA\cite{li2024scemqa} & College-level science MCQs & Targets analytical vs. recall-based questions with structured outputs \\ \hline
CVR\cite{NEURIPS2022_c08ee8fe} & Compositional visual relations & Adds real-world domain grounding and format fidelity metrics \\ \hline
Visual Genome\cite{krishna2017visual} & Dense image annotations & Replaces free-form descriptions with structured analytical tasks \\ \hline
GQA-OOD\cite{acharya2020gqa} & Out-of-distribution VQA & Addresses in-distribution synthetic bias via jury validation \\ \hline
StructMultimodal\cite{gao2020structured} & TextVQA with relation graphs & Generalizes to non-OCR structured data (charts/tables) \\ \hline
ChartQA-X\cite{jang2024chartqa} & Explanations for chart QA & Combines answer+explanation generation in structured formats \\ \hline
SPIQA\cite{smith2024spiqa} & Scientific paper figure QA & Focuses on controlled synthetic data vs. real-world scans \\
\bottomrule
\end{tabular}
\end{table*}

\subsection{Instruction-Following in Vision-Language Models}
Three main strategies have shaped instruction-following in VLMs: (1) task unification (LLaVA \cite{liu2023llava}), which boosts zero-shot performance through broad instruction-response pairings; (2) visual token compression (Qwen-VL \cite{qwen2024}), offering faster inference with minimal accuracy loss; and (3) multi-task alignment (PaLI-X \cite{palix2023}), which reduces forgetting via cross-modal learning. However, instruction-tuned models still lag on multi-hop table reasoning \cite{yang2024enhancing}. GRAFT addresses this gap with chained, multi-step reasoning tasks.

\section{Methodology}
\label{sec:methodology}

\begin{figure*}[!htbp]
\centering
\includegraphics[width=\textwidth, trim=0 0 0 0, clip]{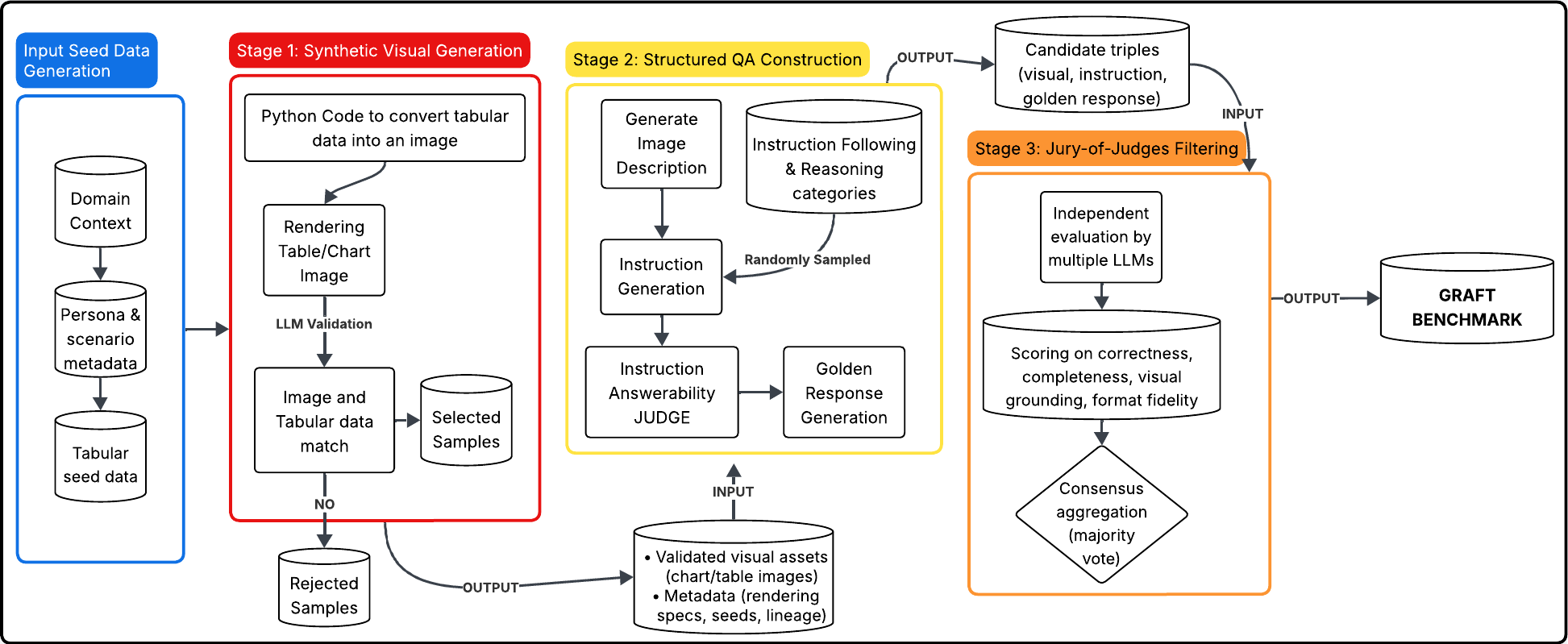}
\caption{GRAFT pipeline for synthetic visual generation and structured reasoning task construction.}
\label{fig:image_generation}
\end{figure*}

We introduce \textbf{GRAFT}, a fully synthetic benchmark pipeline for assessing instruction-following and structured visual reasoning in multimodal large language models (MLLMs) over charts and tables. The pipeline comprises three stages: (\emph{i}) synthetic visual generation, (\emph{ii}) structured QA construction, and (\emph{iii}) multi-model quality control via a jury-of-judges filter. Figure~\ref{fig:image_generation} provides a high-level overview. A complete reference of the variables and notation introduced in this section is provided in Appendix~\ref{app:notation}.

\subsection{Problem Setup and Notation}
\label{subsec:problem_setup}
Let $v$ denote a visual instance (chart or table image), $d(v)$ a structured description of $v$, $q$ a reasoning query grounded in $v$, and $a$ a structured answer adhering to a schema $\mathcal{S}$ (JSON or YAML).
An MLLM $f_\theta$ receives $(v, q)$ and must produce $\hat{a} \in \mathcal{S}$ such that $\hat{a}$ is semantically correct and visually grounded.

\paragraph{Input Spaces.}
Domains $\mathcal{D}$ (e.g., healthcare, finance), personas $\mathcal{P}$, and scenarios $\mathcal{X}$ control content and style; chart types $\mathcal{C}$ and table types $\mathcal{T}$ control structure (Appendices \ref{app:graph-taxonomy}).

\paragraph{Output Targets.}
For each instance we produce $(v, d(v), q, a, \mathsf{meta})$, where $\mathsf{meta}$ stores lineage (random seeds, generators, versioned prompts), and validators’ scores.

\subsection{Stage I: Synthetic Visual Generation}
\label{subsec:visual_generation}

\paragraph{Inputs.}
Domain $d \in \mathcal{D}$, persona $p \in \mathcal{P}$, scenario $x \in \mathcal{X}$; tabular seeds $T \in \mathbb{R}^{m \times n}$ (with $m \in [8,20],\, n \in [3,6]$); chart/table configuration $\kappa$.

\paragraph{Outputs.}
A validated visual $v$ (PNG), its rendering specification $\rho$ (Matplotlib/Seaborn parameters), and metadata $\mathsf{meta}_{\text{viz}}$ (axes labels, legends, color maps, font sizes, DPI, seeds).

\paragraph{Procedure.}
\begin{enumerate}
  \item \textbf{Scenario sampling.} Sample $(d,p,x)$ to bind narrative context and label spaces (units, categories).
  \item \textbf{Data synthesis.} Generate tabular seed $T$ with typed columns (categorical, ordinal, numerical), injected with controllable traits (trend magnitude, noise level, sparsity) via parameterized generators.
  \item \textbf{Chart rendering.} For charts, select $c \in \mathcal{C}$ (bar/line/pie/scatter). Render $v$ via Matplotlib\footnote{\href{https://matplotlib.org/}{https://matplotlib.org/}}/Seaborn\footnote{\href{https://seaborn.pydata.org/}{https://seaborn.pydata.org/}} using $\rho$; enforce axis titles, tick formatting, legends, and numeric precision (Appendix~\ref{app:graph-taxonomy}).
  \item \textbf{Table rendering.} For tables, select $t \in \mathcal{T}$ (pivot, grouped, sortable); format headers, row bands, and cell alignment; render to image with Matplotlib (Appendix \ref{app:table-taxonomy}).
  \item \textbf{LLM-based visual validation.} Use automatic assessors to rate: (i) \emph{type fidelity} (plot/table matches schema), (ii) \emph{label clarity} (axes/headers/legend legible, units present), (iii) \emph{readability} (no occlusion/overplotting). Reject on any failing criterion; otherwise persist $(v,\rho,\mathsf{meta}_{\text{viz}})$.
\end{enumerate}

\subsection{Stage II: Structured QA Construction}
\label{subsec:qa_construction}

\paragraph{Inputs.}
Validated visual $v$ with spec $\rho$; domain context $(d,p,x)$; instruction-type taxonomies $\Gamma_{\text{chart}}$ and $\Gamma_{\text{table}}$ (e.g., ranking, aggregation, comparison, anomaly detection; Appendices \ref{app:graph-instruction-types}); output schema $\mathcal{S}$.

\paragraph{Outputs.}
A tuple $(d(v), q, a)$ where $d(v)$ is a structured description, $q$ is a visual-grounded question answerable solely from $v$, and $a \in \mathcal{S}$ is a machine-parseable answer.

\paragraph{Constraints.}
(i) \emph{No external knowledge:} all facts must be derivable from labels or marks in $v$;
(ii) \emph{Numeric grounding:} computations must reference chart/table labels explicitly;
(iii) \emph{Schema validity:} $a$ must satisfy $\mathcal{S}$ (field names, types, enumerations).

\paragraph{Procedure.}
\begin{enumerate}
  \item \textbf{Saliency extraction.} A VLM summarizes $v$ into $d(v)$ capturing encodings (marks, channels), key trends (slopes, extrema), category relations, and plausible aggregations.
  \item \textbf{Instruction sampling.} Sample a reasoning template $\gamma \in \Gamma_{\text{chart}} \cup \Gamma_{\text{table}}$ conditioned on $(d,p,x)$ and the structure in $d(v)$.
  \item \textbf{Question synthesis.} Instantiate $q$ from $\gamma$ ensuring answerability from $v$ alone; inject clarifying spans (units, time ranges, category subsets) to prevent ambiguity.
  \item \textbf{Answer derivation.} Compute $a$ by programmatic reading of $T$ (the generator’s source of truth) \emph{and} by an independent VLM read from pixels; require agreement. Serialize $a$ to $\mathcal{S}$ (JSON/YAML) with strict formatting.
  \item \textbf{Schema and determinism checks.} Validate $a$ against $\mathcal{S}$; re-run derivation with fixed seeds to ensure deterministic reproduction; reject if mismatched.
\end{enumerate}

\subsection{Stage III: Jury-of-Judges Filtering}
\label{subsec:jury}

\paragraph{Inputs.}
Candidate triples $(v, q, a)$ with $d(v)$ and $\mathsf{meta}$.

\paragraph{Outputs.}
A curated set $\mathcal{G}$ of accepted instances; per-instance diagnostics (scores, rationales, failure tags).

\paragraph{Scoring Dimensions.}
(i) \emph{Correctness} (semantic equivalence to ground truth), (ii) \emph{Completeness} (all requested fields present), (iii) \emph{Visual grounding} (answer aligns to marks/labels in $v$), (iv) \emph{Format fidelity} (valid $\mathcal{S}$) \cite{verga2024replacingjudgesjuriesevaluating}.

\paragraph{Procedure.}
\begin{enumerate}
  \item \textbf{Independent evaluation.} $K$ LLM judges score each $(v,q,a)$ along the four dimensions with calibrated rubrics (Likert or binary per field) and short textual justifications.
  \item \textbf{Consensus aggregation.} Apply majority vote for accept/reject and average for continuous rubrics; tie-break with a deterministic priority (visual grounding $\rightarrow$ correctness $\rightarrow$ format).
  \item \textbf{Adjudication \& repair.} For borderline cases, attempt auto-repair (format-only or minor rounding) under constraints that forbid semantics changes; re-validate schema; otherwise reject.
  \item \textbf{Finalization.} Persist accepted instances to $\mathcal{G}$ with full lineage: $(v,\rho,T,d,p,x,d(v),q,a,\text{scores},\text{judge\_notes},\text{seeds})$.
\end{enumerate}

The GRAFT pipeline combines programmatic data generation, structured reasoning tasks, and rigorous quality control into a unified framework for benchmarking multimodal reasoning. By leveraging synthetic visual assets, schema-constrained QA generation, and multi-model adjudication, GRAFT produces a reproducible and reliable benchmark. Together with the variable reference table in Appendix~\ref{app:notation}, this methodology provides a transparent foundation for evaluating and diagnosing the structured reasoning capabilities of multimodal models.

\section{Experimental Setup}
\label{sec:setup}
For all experimentation and dataset construction presented in this paper, we employed the SyGra\footnote{https://github.com/ServiceNow/SyGra}
 framework \cite{pradhan2025sygra} as the primary tool for data generation. SyGra’s modular, graph-based architecture enables scalable synthesis, quality tagging, and management of multimodal synthetic data, powering the diverse visual–text datasets used throughout our study.

We evaluate vision-language models (VLMs) using the GRAFT benchmark, which is divided into two subsets:

\begin{itemize}
\item \textbf{Chart-QnA}: Reasoning over bar, line, scatter, and pie charts.
\item \textbf{Table-QnA}: Analytical tasks with realistic, domain-specific tables.
\end{itemize}

Each instance provides a rendered chart or table, a natural language question, and a YAML-formatted answer. Tasks include trend analysis, comparison, filtering, aggregation, and multi-step reasoning. Dataset statistics are summarized in Table~\ref{tab:graft_dataset_stats}.

\begin{table*}[!htbp]
    \centering
    \caption{Summary of GRAFT Benchmark Dataset Statistics}
    \label{tab:graft_dataset_stats}
    \begin{tabular}{p{3.5cm} c c c c c}
        \toprule
        \textbf{Benchmark Subset} & \textbf{\#Instances} & \textbf{Avg. Q Tokens} & \textbf{90\% Q Tokens} & \textbf{Avg. A Tokens} & \textbf{90\% A Tokens} \\
        \midrule
        Chart-QnA & 1,412 & 229 & 339 & 164 & 425 \\
        Table-QnA & 1,739 & 323 & 551 & 1,022 & 1,620 \\
        \bottomrule
    \end{tabular}
\end{table*}

Contextual metadata (persona, location, visualization type) is included for each instance, but reasoning must rely solely on the visual input.

\subsection{Benchmarking Pipeline}
\label{subsec:benchmarking_pipeline}

\begin{figure}[!htbp]
\centering
\includegraphics[width=0.5\textwidth]{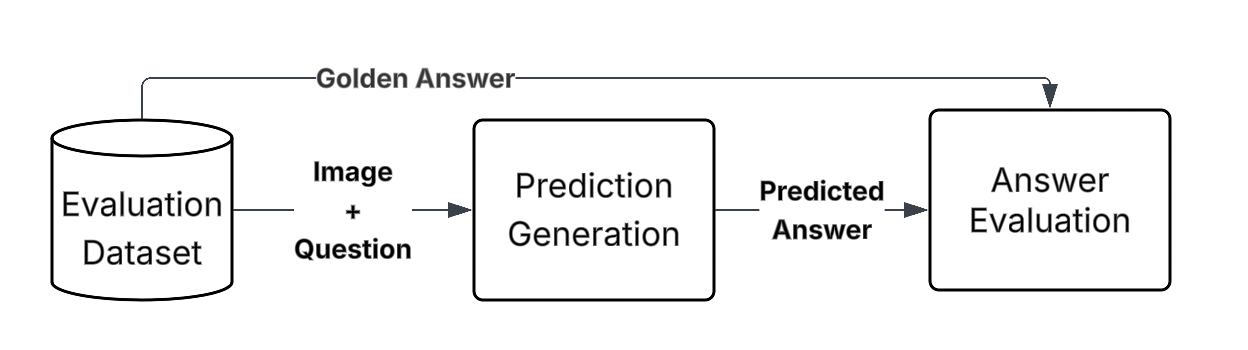}
\caption{Pipeline for Model Evaluation.}
\label{fig:prediction_pipeline}
\end{figure}

We use a two-stage pipeline (see Figure~\ref{fig:prediction_pipeline}):
(1) \textbf{Prediction Generation}: The VLM (e.g., Qwen-2.5-32B-VL) is prompted to generate YAML/JSON answers to each question.
(2) \textbf{Evaluation}: Each answer is scored by a GPT4o-based automated judge.

\subsection{Evaluation Metrics}

Predictions are scored from 1 (poor) to 5 (excellent) along four axes, detailed in Table~\ref{tab:scoring_matrix}:

\begin{itemize}
\item \textbf{Correctness}: Factual alignment with the visual and reference.
\item \textbf{Completeness}: Fully addresses the query.
\item \textbf{Visual Grounding}: Accurate interpretation of visual elements.
\item \textbf{Format Fidelity}: Strict adherence to structured output format.
\end{itemize}

Scores are aggregated to summarize model performance, and the judge provides both numerical scores and brief explanations for deviations.

\section{Discussion}
\label{sec:discussion}

We conduct a qualitative analysis of model performance on GRAFT, focusing on architectural, training, and alignment factors. The goal is to identify where models succeed or fail, and how these outcomes reflect their underlying design choices. We conclude with general takeaways distilled from these observations.

\begin{figure*}[t]
    \centering
    \includegraphics[width=0.70\textwidth]{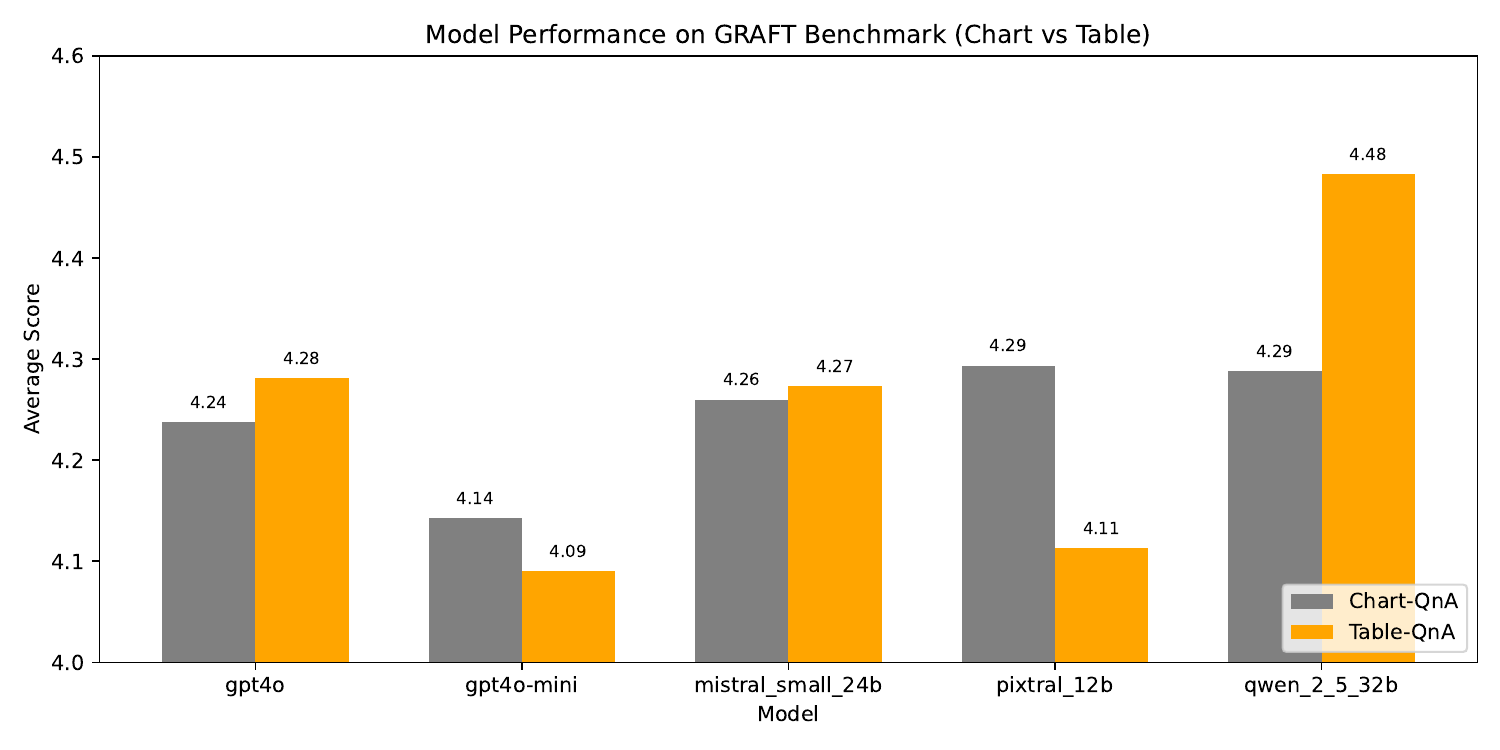}
    \caption{Comparative model performance on Chart-QnA and Table-QnA. Bars represent average scores across all evaluation axes.}
    \label{fig:model-performance-bars}
\end{figure*}

\subsection{Model-Specific Observations}

\textbf{Qwen-2.5 32B VL}\cite{Qwen2.5VL} achieves the most consistent performance across correctness, grounding, and structure. Its strong results on both Chart-QnA and Table-QnA indicate robust reasoning alignment, likely driven by extensive multimodal pretraining and instruction tuning. The stability across modalities suggests well-integrated spatial (chart) and relational (table) representations.

\textbf{Pixtral 12B}\cite{{Pixtral12B}} excels on Chart-QnA, obtaining the highest visual grounding scores, but its accuracy degrades on Table-QnA. This divergence suggests that Pixtral’s strengths lie in shallow visual-text alignment and layout parsing—sufficient for charts—but it lacks the symbolic reasoning depth required for tables. The pattern aligns with training objectives optimized for image-level efficiency rather than structured multi-step reasoning.

\textbf{Mistral-Small-24B-Instruct}\footnote{\href{https://huggingface.co/mistralai/Mistral-Small-24B-Instruct-2501}{https://huggingface.co/mistralai/Mistral-Small-24B-Instruct-2501}} delivers balanced performance across modalities, with particular strength in format fidelity and completeness. Its correctness scores, while solid, lag behind Qwen. The absence of a native visual backbone may explain weaker grounding, with performance likely mediated by cross-modal adaptation layers rather than deeply integrated vision modules.

\textbf{GPT4o}\cite{openai2024gpt4technicalreport} demonstrates high format compliance and completeness, reflecting strong alignment with schema-driven prompting. However, it underperforms on correctness and visual grounding, especially for Chart-QnA. This suggests reliance on OCR-style heuristics rather than genuine visual-semantic parsing, which limits compositional reasoning across visual and textual modalities.

\textbf{GPT4o-mini}, the smaller variant, ranks lowest overall, particularly on correctness and grounding. While it maintains schema fidelity, it struggles with semantic interpretation of structured data. These results illustrate the trade-off between deployment efficiency and reasoning capacity, reinforcing the role of scale in effective multimodal alignment.

\subsection{Key Takeaways}

Our analysis highlights four central findings:

\begin{itemize}
    \item \textbf{Correctness is paramount.} Format fidelity and completeness are insufficient if the underlying reasoning is flawed. Models that consistently produce correct answers should be prioritized over those with superficially higher aggregate scores. 
    
    \item \textbf{Schema adherence is largely solved.} All models reliably produce structured outputs under schema constraints, indicating that prompt-based guidance is effective even for smaller models.
    
    \item \textbf{Visual grounding remains the main bottleneck.} Table-based reasoning in particular exposes weaknesses in capturing implicit relations, hierarchical structures, and long-range dependencies. 
    
    \item \textbf{Instruction tuning and scale drive performance.} Larger, instruction-tuned models (e.g., Qwen, Mistral) achieve better correctness and grounding, underscoring the importance of both capacity and targeted alignment in structured reasoning tasks.
\end{itemize}

Together, these findings outline the current frontier of multimodal reasoning: while structural compliance is achievable, correctness and grounding—especially over relational data—remain unresolved challenges.

\section{Results}
\label{sec:results}

We report model performance on GRAFT using our structured, model-agnostic judging pipeline (Section~\ref{subsec:benchmarking_pipeline}). Each model output is rated 1–5 on \textbf{Correctness}, \textbf{Completeness}, \textbf{Visual Grounding}, and \textbf{Format Fidelity}; scores are averaged per instance and then across the dataset.

\subsection{Quantitative Performance Overview}

Figure~\ref{fig:model-performance-bars} shows mean scores for Chart-QnA and Table-QnA. \textbf{Qwen-2.5 32B VL} leads both subsets. However, average score alone is insufficient—\textbf{correctness} is most crucial, as errors here undermine the value of other metrics. Our analysis thus prioritizes correctness.

For a detailed breakdown of model performance on each subset, see Tables~\ref{tab:table-qna-results} and \ref{tab:chart-qna-results} in the Appendix.

Overall, GRAFT effectively distinguishes between shallow and deep multimodal reasoning abilities. Further discussion is provided in Section \ref{sec:discussion}.

\section{Conclusion}
\label{sec:conclusion}

We introduced the \textbf{GRAFT Benchmark}, a large-scale suite for multimodal question answering over structured tables and charts, designed to test aggregation, comparison, trend analysis, visual grounding, and output fidelity. Using a standardized evaluation pipeline with state-of-the-art vision-language models and LLM-based judging, we found that while models like \textit{GPT4o}, \textit{Pixtral 12B}\footnote{https://huggingface.co/mistralai/Pixtral-12B-2409}, and \textit{Qwen 2.5 32B VL} perform well on format and completeness, they struggle with correctness and visual grounding—especially for complex graphics. These results highlight the need for improved visual reasoning in multimodal systems. GRAFT provides a robust, extensible platform for evaluating and advancing future multimodal QA models.

\section{Limitations and Future Scope}
\label{sec:limitations}
GRAFT has several limitations. First, its reliance on synthetic data may reduce real-world generalizability. Second, language coverage is limited to English, restricting multilingual evaluation. Third, the use of LLM-based judges can sometimes overestimate answer quality, especially in nuanced cases. Finally, some complex chart types—such as heatmaps and sankey diagrams—are underrepresented in the current version.

Future improvements include incorporating human-annotated benchmarks, real-world and multilingual data, richer task formats (e.g., multi-hop reasoning), and finer-grained evaluation of visual grounding. These enhancements will help GRAFT remain a robust and evolving multimodal benchmark.



\onecolumn
\appendix
\section*{Notation and Variable References}
\label{app:notation}

Table \ref{tab:notation} summarizes the mathematical notation and variables introduced in Section \ref{sec:methodology}.

\begin{table*}[!htbp]
\centering
\caption{Reference table of symbols and variables used in the GRAFT methodology.}
\label{tab:notation}
\begin{tabular}{p{2.5cm} p{12cm}}
\toprule
\textbf{Symbol / Variable} & \textbf{Description} \\
\midrule
$v$ & A visual instance, either a chart or a table image. \\
$d(v)$ & Structured description of visual instance $v$, generated by a vision--language model (VLM). \\
$q$ & A reasoning query grounded in $v$, constructed from instruction templates. \\
$a$ & Ground-truth structured answer to $q$, expressed in schema $\mathcal{S}$ (JSON/YAML). \\
$\hat{a}$ & Model-predicted answer produced by an MLLM $f_\theta$. \\
$f_\theta$ & A multimodal large language model (MLLM) parameterized by $\theta$, mapping $(v, q) \mapsto \hat{a}$. \\
$\mathcal{S}$ & Structured output schema (e.g., JSON or YAML), enforcing field and type constraints. \\
\midrule
$\mathcal{D}$ & Domain space (e.g., healthcare, finance, education). \\
$\mathcal{P}$ & Persona space, defining the role or perspective in which queries are posed. \\
$\mathcal{X}$ & Scenario space, specifying contextual settings associated with each instance. \\
$\mathcal{C}$ & Chart type space (bar, line, pie, scatter). \\
$\mathcal{T}$ & Table type space (pivot, grouped, sortable). \\
\midrule
$d, p, x$ & Individual sampled domain, persona, and scenario, respectively. \\
$T \in \mathbb{R}^{m \times n}$ & Tabular seed data with $m$ rows ($8 \leq m \leq 20$) and $n$ columns ($3 \leq n \leq 6$). \\
$c \in \mathcal{C}$ & Specific chart type selected for rendering. \\
$t \in \mathcal{T}$ & Specific table type selected for rendering. \\
$\kappa$ & Chart/table rendering configuration (layout and style). \\
$\rho$ & Rendering specification (Matplotlib/Seaborn parameters: axes, labels, colors, legends, DPI, etc.). \\
$\mathsf{meta}_{\text{viz}}$ & Metadata associated with a rendered visual, including seeds, rendering parameters, and validators’ scores. \\
\midrule
$\Gamma_{\text{chart}}$ & Instruction taxonomy for chart-based reasoning tasks. \\
$\Gamma_{\text{table}}$ & Instruction taxonomy for table-based reasoning tasks. \\
$\gamma$ & A specific reasoning template sampled from $\Gamma_{\text{chart}} \cup \Gamma_{\text{table}}$. \\
\midrule
$\mathcal{G}$ & Final curated benchmark set of accepted instances after jury-of-judges filtering. \\
$K$ & Number of LLMs acting as independent judges. \\
\text{scores} & Evaluation metrics assigned by judges (correctness, completeness, grounding, format). \\
\text{judge\_notes} & Explanatory rationales provided by judges during evaluation. \\
\text{seeds} & Random seeds used in data synthesis, rendering, and validation for reproducibility. \\
\bottomrule
\end{tabular}
\end{table*}

\section*{Instruction Categories}
\label{sec:instruction_categories}

This appendix provides an overview of the instruction categories used within GRAFT to generate a broad range of question types across graphs, tables, natural images, and posters.

\subsection*{Graph Generation Taxonomy}
\label{app:graph-taxonomy}

Graph-focused questions in GRAFT draw on high-quality, programmatically rendered charts derived from synthetic, domain-specific datasets. The generation pipeline is designed to produce visually varied plots with realistic semantics and contextually grounded personas. Table \ref{app:table-taxonomy} summarizes the major dimensions that shape chart synthesis.

\begin{table}[!htbp]
\centering
\caption{Taxonomy of Graph Image Generation Attributes}
\label{tab:graph_generation_taxonomy}
\begin{tabular}{p{3cm} p{14cm}}
\toprule
\textbf{Dimension} & \textbf{Description and Examples} \\
\midrule
\textbf{Domain Genre} & Selected from over 25 application areas, such as Finance, Education, Environment, and Cybersecurity. Each genre is paired with realistic user roles—researchers, educators, analysts, policymakers—to situate charts in plausible scenarios. \\
    \textbf{Geographic Context} & Drawn from a pool of more than 50 global cities and countries, supporting culturally grounded and region-specific contexts (e.g., Nairobi, Tokyo, San Francisco, Dubai). \\

    \textbf{Plot Type} & Determines the visual encoding and structural form of the chart. Supported types include:
    \begin{itemize}
        \item \texttt{line}, \texttt{bar}, \texttt{scatter}, \texttt{hist}, \texttt{kde}
        \item \texttt{box}, \texttt{heatmap}, \texttt{pie}, \texttt{area}, \texttt{violin}
        \item \texttt{count}, \texttt{strip}, \texttt{swarm}
    \end{itemize}
    Plot selection is guided by the semantics of the underlying data—for instance, using line charts for temporal trends or box plots for distributional comparisons. \\

    \textbf{Visual Styling} & Charts are rendered using Python libraries such as Matplotlib and Seaborn, with systematic variation in:
    \begin{itemize}
        \item Axis label design, tick density, and color schemes
        \item Title fonts and spacing
        \item DPI scaling (150–350) to introduce variation in clarity and resolution
    \end{itemize} \\

    \textbf{Semantic Context} & Each chart is embedded within a structured metadata schema capturing:
    \begin{itemize}
        \item Persona (e.g., financial analyst, high school teacher)
        \item Scenario (e.g., quarterly earnings review, academic performance summary)
        \item Timeframe and geographic references
    \end{itemize} \\
    \bottomrule
\end{tabular}
\end{table}

Together, these dimensions create a diverse and semantically rich graph dataset that supports a wide spectrum of reasoning tasks grounded in realistic visual contexts.

\subsection*{Graph Instruction Categories}
\label{app:graph-instruction-types}

Questions involving graphs are designed to test fundamental visual reasoning capabilities, each targeting a specific analytic skill. Importantly, all questions rely solely on the information conveyed by the chart itself.

\begin{table*}[!htbp]
\centering
\caption{Instruction Taxonomy for Graph-Based QA}
\label{tab:graph_categories}
\begin{tabular}{p{2.5cm} p{6cm} p{7cm}}
\toprule
\textbf{Category} & \textbf{Description} & \textbf{Example Question} \\
\midrule
Comparison & Assess differences or similarities across groups or categories & Which group had the highest value? \\
Trend Identification & Detect directional patterns, slopes, or long-term changes & What trend is observed from 2010 to 2020? \\
Change Calculation & Compute changes between points or over time & By how much did revenue grow from Q1 to Q2? \\
Ranking & Order categories by magnitude or intensity & Rank regions by number of cases reported. \\
Proportion/Share & Evaluate segment sizes or relative contributions & What share of traffic is from mobile users? \\
Outlier Detection & Identify visually anomalous or unexpected points & Which year shows an unusual spike in complaints? \\
Inference & Perform multi-step or projection-based reasoning & If the trend continues, when will the value exceed 1{,}000? \\
Aggregation & Combine values across series or categories & What is the combined total of Category A and B? \\
Ratio/Rate & Infer ratios or per-unit metrics from visual encodings & What is the dropout rate per 100 students? \\
\bottomrule
\end{tabular}
\end{table*}

This taxonomy ensures structured coverage of the reasoning operations commonly required for interpreting data visualizations, enabling precise benchmarking of model performance on graph-based tasks.

\subsection*{Table Generation Taxonomy}
\label{app:table-taxonomy}

To enable rigorous table-based reasoning, we define a detailed taxonomy governing how synthetic tables are generated. This encompasses semantic content, structural layout, visual presentation, and contextual grounding. Table \ref{tab:table_generation_taxonomy} outlines the primary dimensions of variation.

\begin{table}[!htbp]
\centering
\caption{Taxonomy of Table Image Generation Attributes}
\label{tab:table_generation_taxonomy}
\begin{tabular}{p{3cm} p{14cm}}
\toprule
\textbf{Dimension} & \textbf{Description and Examples} \\
\midrule
\textbf{Domain Genre} & Selected from over 25 areas—including Healthcare, Finance, and Technology—each paired with personas such as doctors, analysts, teachers, or urban planners. \\
    \textbf{Geographic Context} & Includes real-world locations (e.g., Tokyo, Cape Town, London) for greater authenticity and situational grounding. \\

    \textbf{Table Type} & Defines structural layout and data organization, including:
    \begin{itemize}
        \item \texttt{flat}, \texttt{grouped}, \texttt{pivot}, \texttt{sortable}, \texttt{highlighted}
        \item \texttt{comparison}, \texttt{ranked}, \texttt{time\_series}, \texttt{proportion}, \texttt{matrix}
    \end{itemize} \\

    \textbf{Visual Styling} & Controlled variation in presentation, including:
    \begin{itemize}
        \item Font families (e.g., Arial, STIXGeneral), sizes (8–14pt)
        \item Header/row color schemes, border styles
        \item Row/column scaling, padding, zebra striping
        \item Render DPI (150–350)
    \end{itemize} \\

    \textbf{Layout Controls} & Additional parameters governing:
    \begin{itemize}
        \item Minimum table dimensions
        \item Title fonts and spacing
        \item Highlighting logic for key cells
    \end{itemize} \\
    \bottomrule
\end{tabular}
\end{table}

These dimensions ensure that generated tables vary meaningfully in structure, semantics, and visual style—mirroring the diversity seen in real-world dashboards and analytic interfaces.

\subsection*{Table Instruction Categories}
\label{app:table-instruction-types}

Table-based questions target a broad set of analytical skills needed for interpreting structured visual data. The following taxonomy defines the core categories used throughout GRAFT, each illustrated with representative question types.

\begin{table}[!htbp]
\centering
\caption{Instruction Taxonomy for Table-Based QA}
\label{tab:table_categories}
\begin{tabular}{p{3cm} p{5.5cm} p{6.5cm}}
\toprule
\textbf{Category} & \textbf{Description} & \textbf{Example Question} \\
\midrule
Comparison & Contrast values across rows, columns, or groups & Which group has the highest value? \\
Pattern Recognition & Identify recurring numerical or structural patterns & Which column shows a consistent increase? \\
Change Calculation & Measure shifts over time or between categories & What changed most between Year X and Y? \\
Ranking & Order entries based on numeric criteria & Rank departments by average score. \\
Proportion/Share & Compute part-to-whole or percentage-based relations & What \% of revenue comes from Product A? \\
Anomaly Detection & Flag unusually high or low values & Which month saw an unexpected drop? \\
Deductive Inference & Consider multi-step or hypothetical scenarios & If salaries rise by 10\%, what's the total cost? \\
Aggregation & Combine values via summation or averaging & What is the total enrollment in Q3? \\
Ratio/Rate & Derive ratios, per-capita measures, or unit rates & What is the conversion rate per 100 visits? \\
\bottomrule
\end{tabular}
\end{table}

Together, these categories offer comprehensive coverage of the reasoning operations that underlie structured data interpretation, supporting detailed evaluation of models’ visual understanding and analytical capabilities.

\section*{Detailed Model Results}
\label{sec:appendix-detailed-tables}

\subsection*{Table-QnA Results}

Before presenting the results, it is helpful to highlight that the \textbf{Table-QnA} portion of GRAFT (1,739 examples) is intentionally designed to stress structured tabular reasoning. Tasks in this subset place particular weight on factual accuracy and internal consistency, as well as on adherence to the required output format. Each model is evaluated along several complementary dimensions—correctness, completeness, visual grounding, and format fidelity—to capture both its reasoning quality and its ability to follow structured instructions.

\begin{table}[H]
\centering
\caption{Evaluation scores on the \textbf{Table-QnA} subset of GRAFT (1,739 examples).}
\label{tab:table-qna-results}
\small
\begin{tabular}{lccccc}
\toprule
\textbf{Model} & \textbf{Correctness} & \textbf{Completeness} & \textbf{Visual Grounding} & \textbf{Format Fidelity} & \textbf{Avg. Score} \\
\midrule
Qwen 2.5 32B VL      & \textbf{4.20} & 4.90 & \textbf{3.85} & 4.98 & \textbf{4.48} \\
GPT4o                 & 3.78 & \textbf{4.91} & 3.46 & 4.97 & 4.28 \\
Mistral 24B           & 3.76 & \textbf{4.91} & 3.44 & \textbf{4.98} & 4.27 \\
Pixtral 12B           & 3.40 & 4.88 & 3.17 & \textbf{4.99} & 4.11 \\
GPT4o-mini            & 3.39 & 4.80 & 3.20 & 4.97 & 4.09 \\
\bottomrule
\end{tabular}
\end{table}

As summarized in Table \ref{tab:table-qna-results}, Qwen 2.5 32B VL delivers the strongest overall performance on this subset, with notably high scores in correctness and grounding. GPT4o and Mistral 24B follow closely, performing reliably across all metrics. In contrast, smaller models such as GPT4o-mini and Pixtral 12B maintain good format fidelity yet fall short in correctness and grounding, reflecting the limitations commonly associated with reduced parameter capacity.

\subsection*{Chart-QnA Results}

The \textbf{Chart-QnA} subset (1,412 examples) shifts the emphasis toward interpreting quantitative information encoded in charts. These tasks require models to blend numerical reasoning with careful reading of visual structure, making grounding accuracy especially important. The same evaluation criteria used for Table-QnA apply here as well.

\begin{table}[H]
\centering
\caption{Evaluation scores on the \textbf{Chart-QnA} subset of GRAFT (1,412 examples).}
\label{tab:chart-qna-results}
\small
\begin{tabular}{lccccc}
\toprule
\textbf{Model} & \textbf{Correctness} & \textbf{Completeness} & \textbf{Visual Grounding} & \textbf{Format Fidelity} & \textbf{Avg. Score} \\
\midrule
Pixtral 12B          & \textbf{3.92} & 4.90 & \textbf{3.37} & \textbf{4.99} & \textbf{4.29} \\
Qwen 2.5 32B VL      & \textbf{3.92} & 4.90 & 3.34 & \textbf{4.99} & \textbf{4.29} \\
Mistral 24B          & 3.86 & \textbf{4.94} & 3.25 & \textbf{4.99} & 4.26 \\
GPT4o                & 3.83 & 4.88 & 3.25 & \textbf{4.99} & 4.24 \\
GPT4o-mini           & 3.62 & 4.87 & 3.08 & \textbf{4.99} & 4.14 \\
\bottomrule
\end{tabular}
\end{table}

According to the results presented in Table \ref{tab:chart-qna-results}, Pixtral 12B and Qwen 2.5 32B VL achieve the top average scores, driven by their consistently strong correctness and grounding across diverse chart types. Mistral 24B and GPT4o are competitive, while GPT4o-mini performs well but falls short relative to larger counterparts.

\subsection*{Procedural Overview of Extended Analysis}

To further interpret why certain architectures perform differently, we conducted a structured post-hoc analysis using quantitative correlations and qualitative inspection.  
The procedure followed these steps:

\begin{enumerate}
    \item \textbf{Data aggregation:} We combined model-level metrics from Tables\ref{tab:table-qna-results} and\ref{tab:chart-qna-results} into a single structured dataset. Each row represented a (\textit{model, subset}) pair with correctness, grounding, completeness, format fidelity, and average score.
    \item \textbf{Model metadata annotation:} For each model, we manually assigned three categorical attributes — \emph{architecture type} (fusion, compression, adapter), \emph{instruction-tuning status}, and approximate \emph{parameter size (in billions)} — based on public model documentation.
    \item \textbf{Cross-subset averaging:} To mitigate sample-size imbalance between Chart-QnA and Table-QnA, we computed per-model averages across both subsets.
    \item \textbf{Modality sensitivity computation:} We calculated $\Delta$ (Table–Chart) differences for each metric, indicating whether models gained or lost accuracy when transitioning from structured to visual data.
    \item \textbf{Correlation analysis:} We then computed Pearson correlations between numeric metadata (e.g., parameter size, architecture dummies) and averaged performance metrics. Although the sample size is limited ($n=5$), the analysis highlights trends between architectural design and reasoning performance.
    \item \textbf{Visualization and interpretation:} Results were visualized using bar charts and scatter plots (not shown here for brevity) to cross-check consistency between numeric trends and qualitative observations.
\end{enumerate}

\subsection*{Expanded Analysis: Architectural and Training Factors}

Beyond descriptive trends, we analyze why specific architectures succeed or fail on GRAFT. Models with \emph{multimodal fusion} encoders (e.g., Qwen-2.5-32B-VL) show higher \emph{Correctness} and \emph{Visual Grounding} on average than models emphasizing \emph{vision-token compression} (e.g., Pixtral 12B). Fusion enables deeper cross-modal interaction and fine-grained alignment between marks and tokens, which is critical for table reasoning. Instruction-tuned models also exhibit stronger schema adherence and fewer hallucinations. Finally, exposure to analytical chart/table data during pretraining correlates with improved grounding, indicating that text-heavy corpora alone are insufficient for structured visual reasoning.

\begin{table}[H]
\centering
\caption{Model metadata used in the interpretive analysis.}
\label{tab:model-metadata}
\small
\begin{tabular}{lccc}
\toprule
\textbf{Model} & \textbf{Architecture} & \textbf{Instruction-tuned} & \textbf{Params (B)} \\
\midrule
Qwen 2.5 32B VL  & fusion       & Yes & 32 \\
GPT4o            & fusion       & Yes & --- \\
Mistral 24B      & adapter      & Yes & 24 \\
Pixtral 12B      & compression  & Yes & 12 \\
GPT4o-mini       & fusion       & Yes & --- \\
\bottomrule
\end{tabular}
\end{table}

\begin{table}[H]
\centering
\caption{Per-model averages across both subsets (Chart \& Table).}
\label{tab:per-model-avg}
\small
\begin{tabular}{lccccc}
\toprule
\textbf{Model} & \textbf{Correctness} & \textbf{Grounding} & \textbf{Completeness} & \textbf{Format} & \textbf{Avg. Score} \\
\midrule
Qwen 2.5 32B VL  & \textbf{4.06} & \textbf{3.60} & 4.90 & 4.99 & \textbf{4.39} \\
Mistral 24B      & 3.81 & 3.35 & \textbf{4.93} & 4.99 & 4.27 \\
GPT4o            & 3.81 & 3.36 & 4.90 & 4.98 & 4.26 \\
Pixtral 12B      & 3.66 & 3.27 & 4.89 & \textbf{4.99} & 4.20 \\
GPT4o-mini       & 3.51 & 3.14 & 4.84 & 4.98 & 4.12 \\
\bottomrule
\end{tabular}
\end{table}

\begin{table}[H]
\centering
\caption{Modality sensitivity: $\Delta$ (Table–Chart) for each metric. Positive values mean Table-QnA $>$ Chart-QnA.}
\label{tab:modality-deltas}
\small
\begin{tabular}{lccccc}
\toprule
\textbf{Model} & $\Delta$ \textbf{Correct.} & $\Delta$ \textbf{Ground.} & $\Delta$ \textbf{Complete.} & $\Delta$ \textbf{Format} & $\Delta$ \textbf{Avg.} \\
\midrule
Qwen 2.5 32B VL  & \textbf{+0.28} & \textbf{+0.26} & 0.00 & -0.01 & \textbf{+0.19} \\
Mistral 24B      & -0.10 & +0.09 & -0.03 & -0.01 & +0.01 \\
GPT4o            & -0.05 & +0.20 & +0.03 & -0.02 & +0.04 \\
GPT4o-mini       & -0.23 & -0.04 & -0.07 & -0.02 & -0.05 \\
Pixtral 12B      & -0.52 & -0.20 & -0.02 & 0.00 & -0.18 \\
\bottomrule
\end{tabular}
\end{table}

\begin{table}[H]
\centering
\caption{Exploratory Pearson correlations (averaged across subsets).}
\label{tab:correlations}
\small
\begin{tabular}{lccc}
\toprule
\textbf{Target Metric} & \textbf{Params (B)} & \textbf{Arch: Compression} & \textbf{Arch: Fusion} \\
\midrule
Correctness     & \phantom{-}0.967 & -0.294 & \phantom{-}0.146 \\
Visual Grounding& \phantom{-}0.915 & -0.239 & \phantom{-}0.184 \\
Completeness    & \phantom{-}0.386 & \phantom{-}0.017 & -0.511 \\
Format Fidelity & -0.918           & \phantom{-}0.802 & -0.764 \\
Avg. Score      & \phantom{-}0.960 & -0.254 & \phantom{-}0.115 \\
\bottomrule
\end{tabular}
\end{table}

(1) Larger models tend to correlate positively with \emph{Correctness} and \emph{Grounding}, but not with \emph{Format}, which is uniformly high. (2) The negative association of \texttt{compression} with correctness/grounding and the positive association of \texttt{fusion} are consistent with our qualitative analysis that deeper cross-modal fusion benefits structured visual reasoning. (3) Completeness and format appear largely saturated across models, suggesting that future improvements should prioritize semantic grounding and factual accuracy.

\section*{Evaluation Criteria}
\label{sec:evaluation_criteria_appendix}

To ensure consistent, transparent, and reproducible assessment of model responses in GRAFT, we employ a structured scoring matrix evaluated by multiple independent LLM judges (see Section \ref{subsec:jury}). Each response is rated along four complementary axes—\textbf{Correctness}, \textbf{Completeness}, \textbf{Visual Grounding}, and \textbf{Format Fidelity}—using a calibrated 1–5 Likert scale. 

Prior to scoring, all judges undergo a short calibration phase using gold-standard examples with annotated rationales to harmonize their interpretation of scale anchors. Inputs are presented in randomized order, with no shared memory across evaluations, ensuring independence of judgments. Aggregated scores are derived via majority voting for discrete metrics and mean pooling for continuous ones.  
To confirm the stability and independence of the evaluation metrics, inter-rater agreement was computed on a stratified subset of 50 instances, yielding Cohen’s~$\kappa \approx 0.81$, which reflects substantial consistency across judges and minimal correlated bias.

This multi-judge, bias-controlled framework enables fine-grained, aspect-based evaluation and supports robust comparative analysis of structured reasoning capabilities in multimodal models.

\begin{table*}[!htbp]
    \centering
    \caption{Scoring Matrix for Evaluation Metrics (Scale: 1–5). Judges were calibrated on reference examples to ensure consistent interpretation of scale anchors.}
    \label{tab:scoring_matrix}
    \begin{tabular}{p{4cm} p{1cm} p{9cm}}
        \toprule
        \textbf{Evaluation Axis} & \textbf{Score} & \textbf{Interpretation} \\
        \midrule

        \multirow{5}{*}{\textbf{Correctness}} 
        & 5 & Fully correct and factually aligned with chart/table and reference. \\
        & 4 & One minor factual error with negligible impact. \\
        & 3 & Moderate inaccuracies that weaken reliability. \\
        & 2 & Major factual mistake affecting the core content. \\
        & 1 & Entirely incorrect answer. \\

        \midrule

        \multirow{5}{*}{\textbf{Completeness}} 
        & 5 & Completely answers all parts of the question. \\
        & 4 & One small aspect is missing. \\
        & 3 & Several required components are missing. \\
        & 2 & Only partially answers the question. \\
        & 1 & Largely incomplete or missing response. \\

        \midrule

        \multirow{5}{*}{\textbf{Visual Grounding}} 
        & 5 & Perfect interpretation of visual elements (e.g., chart marks, labels, or cell references). \\
        & 4 & Slight visual misread but generally correct. \\
        & 3 & Some visual elements misinterpreted or ambiguously mapped. \\
        & 2 & Several major misreads of visual content. \\
        & 1 & No clear connection between the answer and visual data. \\

        \midrule

        \multirow{5}{*}{\textbf{Format Fidelity}} 
        & 5 & Output matches schema and formatting exactly. \\
        & 4 & Minor format issues (e.g., spacing, punctuation, casing). \\
        & 3 & Multiple format errors, still syntactically valid. \\
        & 2 & Major structure breakdown (e.g., missing fields or invalid keys). \\
        & 1 & Completely incorrect or unusable format. \\

        \bottomrule
    \end{tabular}
\end{table*}

\section*{Dataset Preview}
\label{appendix:dataset-preview}

The following appendix provides a visual walkthrough of sample records drawn from the GRAFT benchmark. Each record corresponds to a visual reasoning QA instance—grounded in either a chart or a table—and includes both the visual input and detailed annotations used during training and evaluation.

Each data record is composed of the following key fields:

\begin{itemize}
    \item \textbf{id:} A unique SHA-256 hash representing the data instance.
    
    \item \textbf{image:} The rendered chart or table image used as the sole visual input to the model.
    
    \item \textbf{conversation:} A structured list capturing the instruction-response exchange. This includes the visual reference (image) followed by a complex, multimodal question and a YAML-structured answer generated by a VLM.
    
    \item \textbf{metadata:} Encapsulates the quality and validity judgments for each record, including:
    \begin{itemize}
        \item \texttt{answerability.JUDGEMENT}: Whether the question is visually answerable.
        \item \texttt{answerability.JUDGEMENT \_EXPLANATION}: Explanation for the answerability tag.
        \item \texttt{correct\_answer\_evaluation.JUDGEMENT}: Whether the answer is factually and visually correct.
        \item \texttt{chosen\_formattedness.JUDGEMENT}: Assessment of whether the answer adheres to the expected YAML/JSON format.
        \item \texttt{generated\_instruction},  \texttt{generated\_answer}: The actual model-generated instruction and structured output.
        \item \texttt{image\_description}: Auto-generated natural language description of the visual, capturing axes, legends, trends, and key patterns.
    \end{itemize}
    
    \item \textbf{judgement\_metadata:} A consolidated set of scores and explanations from multiple model-based judges (e.g., GPT-4o, Pixtral, Mistral), including their independent verdicts and rationales. This field helps support ensemble-based QA validation pipelines described in Section \ref{subsec:benchmarking_pipeline}.

\end{itemize}

Each previewed record in the pages below is presented as a compact tabular summary with its associated chart or table image. Fields with long content are truncated for readability (indicated by ellipses).

\vspace{1em}

\noindent\textit{Sample rendered records are shown below for the first five examples in the benchmark:}

\clearpage
\thispagestyle{empty}
\begin{figure}[p]
  \centering
  \includegraphics[
    width=\dimexpr\paperwidth-4cm\relax,
    keepaspectratio,
    trim=2cm 1cm 1cm 1cm, clip
  ]{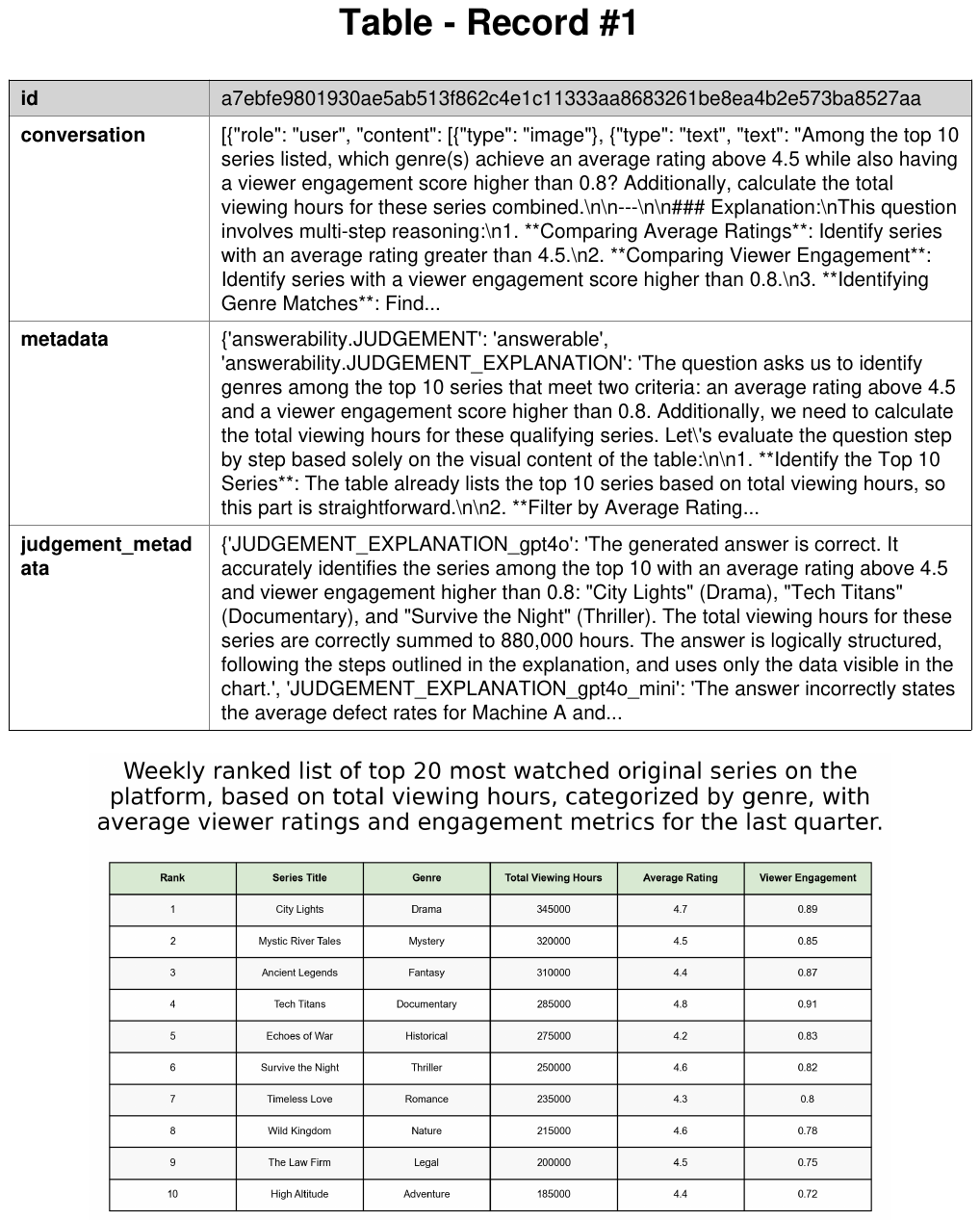}
\end{figure}

\clearpage
\thispagestyle{empty}
\begin{figure}[p]
  \centering
  \includegraphics[
    width=\dimexpr\paperwidth-4cm\relax,
    keepaspectratio,
    trim=2cm 1cm 1cm 1cm, clip
  ]{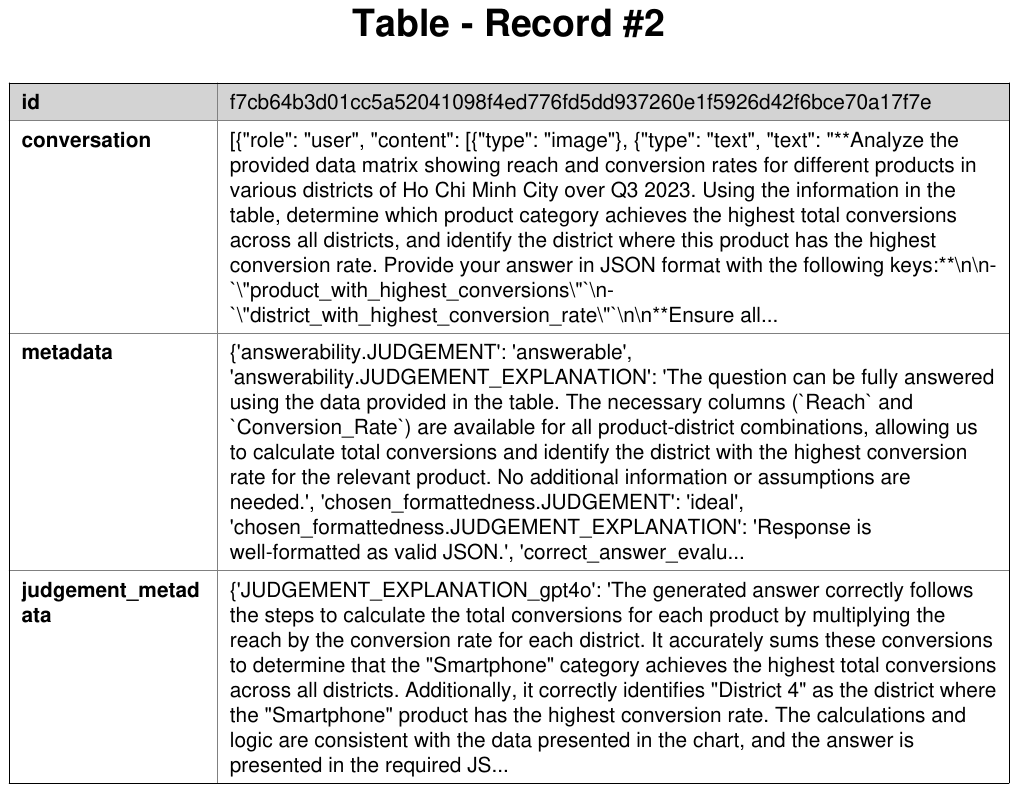}
\end{figure}

\clearpage
\thispagestyle{empty}
\begin{figure}[p]
  \centering
  \includegraphics[
    width=\dimexpr\paperwidth-4cm\relax,
    keepaspectratio,
    trim=2cm 1cm 1cm 1cm, clip
  ]{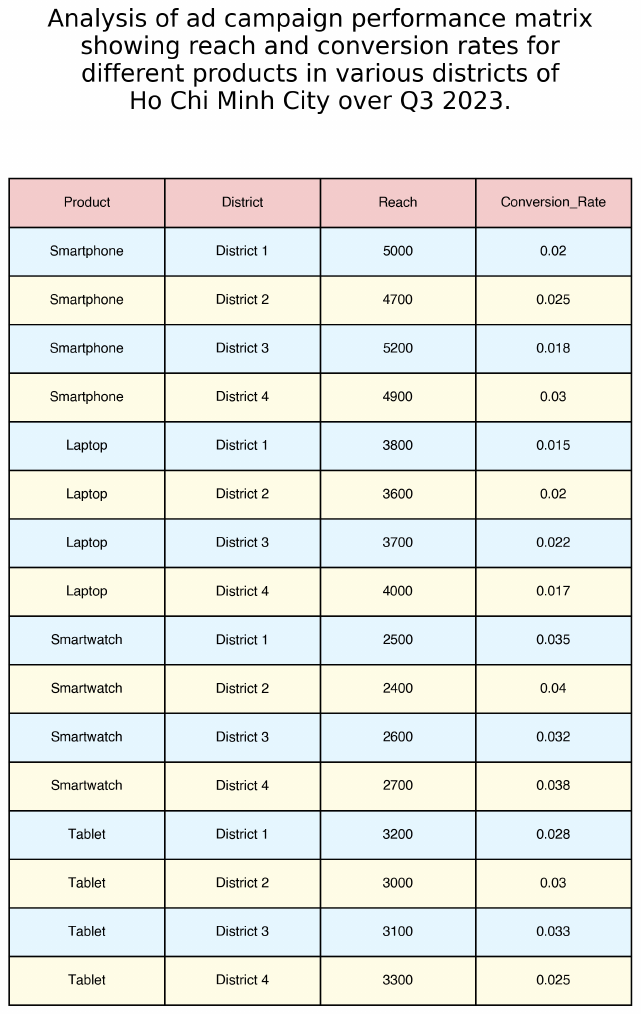}
\end{figure}

\clearpage
\thispagestyle{empty}
\begin{figure}[p]
  \centering
  \includegraphics[
    width=\dimexpr\paperwidth-4cm\relax,
    keepaspectratio,
    trim=2cm 1cm 1cm 1cm, clip
  ]{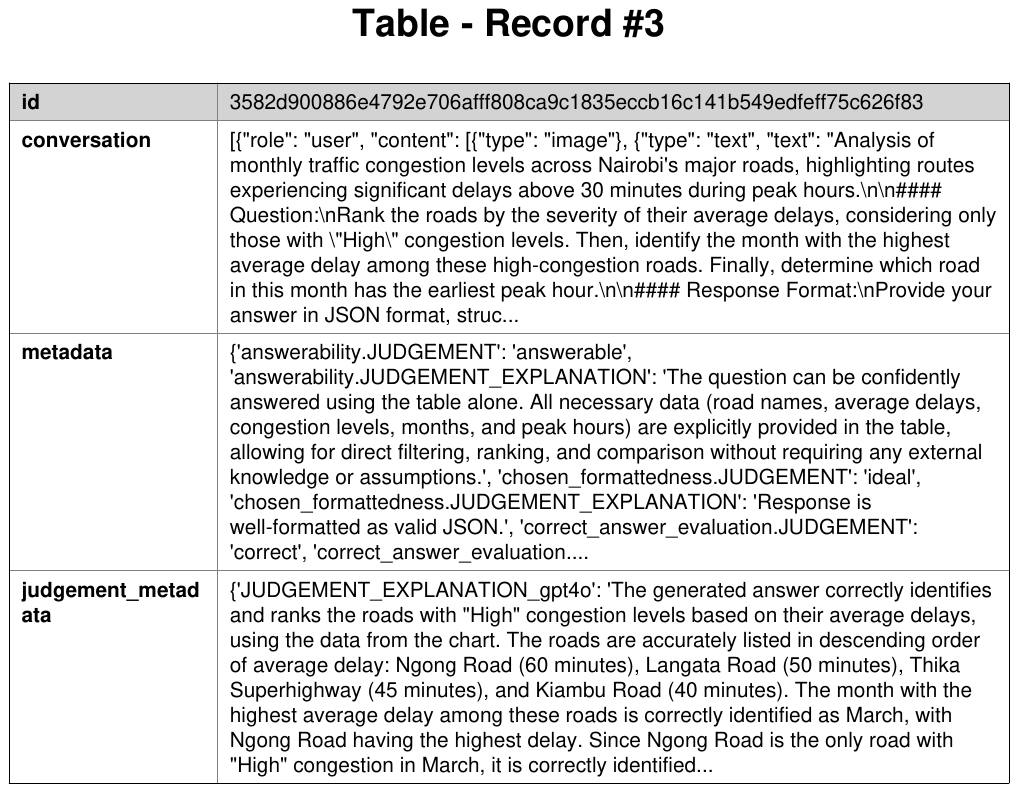}
\end{figure}

\clearpage
\thispagestyle{empty}
\begin{figure}[p]
  \centering
  \includegraphics[
    width=\dimexpr\paperwidth-4cm\relax,
    keepaspectratio,
    trim=2cm 1cm 1cm 1cm, clip
  ]{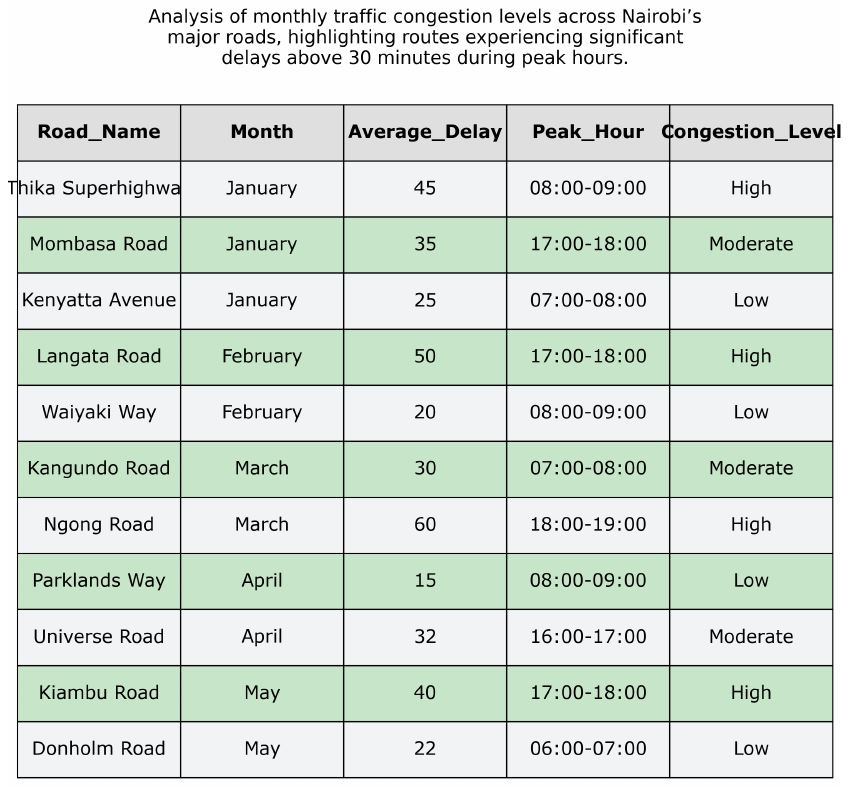}
\end{figure}

\clearpage
\thispagestyle{empty}
\begin{figure}[p]
  \centering
  \includegraphics[
    width=\dimexpr\paperwidth-4cm\relax,
    keepaspectratio,
    trim=2cm 1cm 1cm 1cm, clip
  ]{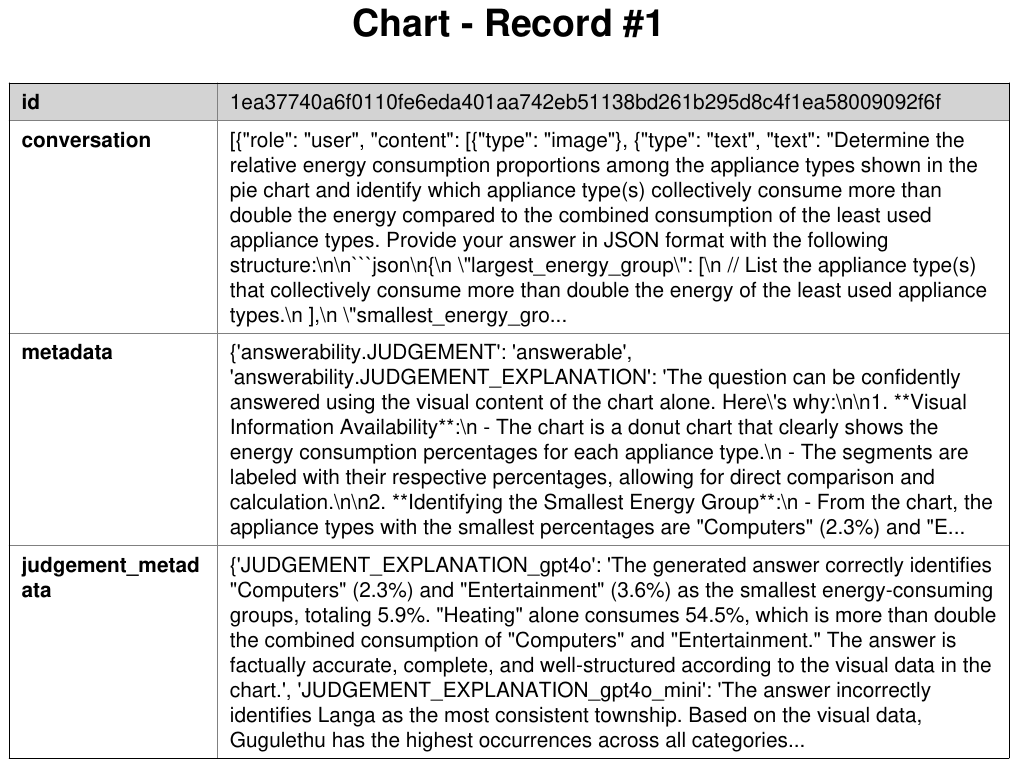}
\end{figure}

\clearpage
\thispagestyle{empty}
\begin{figure}[p]
  \centering
  \includegraphics[
    width=\dimexpr\paperwidth-4cm\relax,
    keepaspectratio,
    trim=2cm 1cm 1cm 1cm, clip
  ]{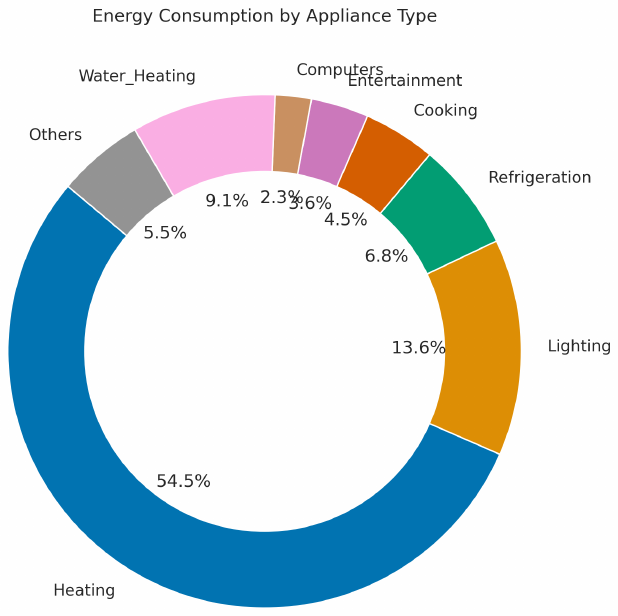}
\end{figure}

\clearpage
\thispagestyle{empty}
\begin{figure}[p]
  \centering
  \includegraphics[
    width=\dimexpr\paperwidth-4cm\relax,
    keepaspectratio,
    trim=2cm 1cm 1cm 1cm, clip
  ]{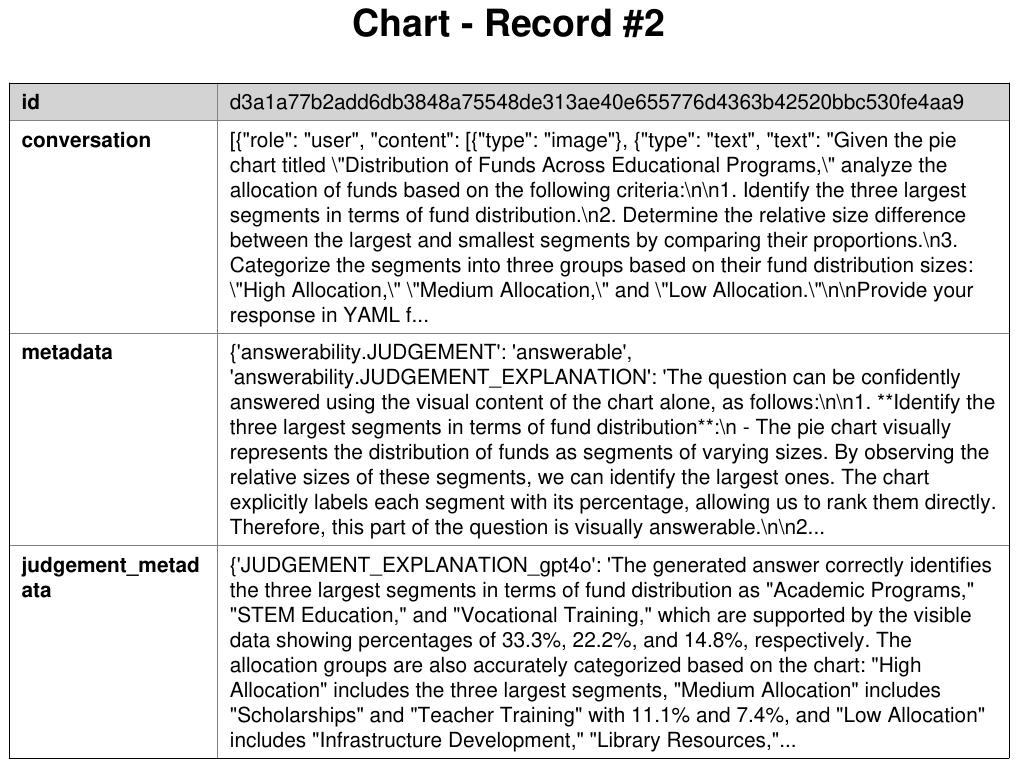}
\end{figure}

\clearpage
\thispagestyle{empty}
\begin{figure}[p]
  \centering
  \includegraphics[
    width=\dimexpr\paperwidth-4cm\relax,
    keepaspectratio,
    trim=2cm 1cm 1cm 1cm, clip
  ]{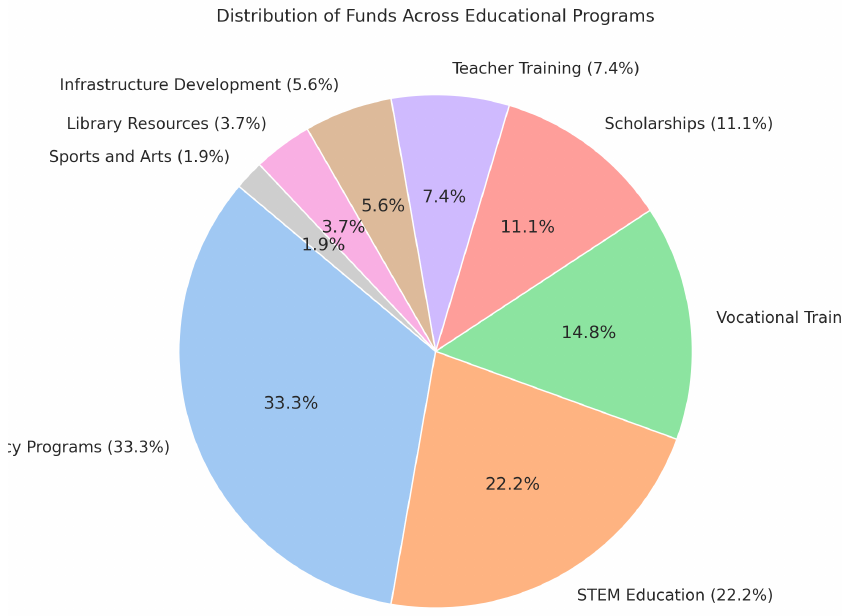}
\end{figure}

\clearpage
\thispagestyle{empty}
\begin{figure}[p]
  \centering
  \includegraphics[
    width=\dimexpr\paperwidth-4cm\relax,
    keepaspectratio,
    trim=2cm 1cm 1cm 1cm, clip
  ]{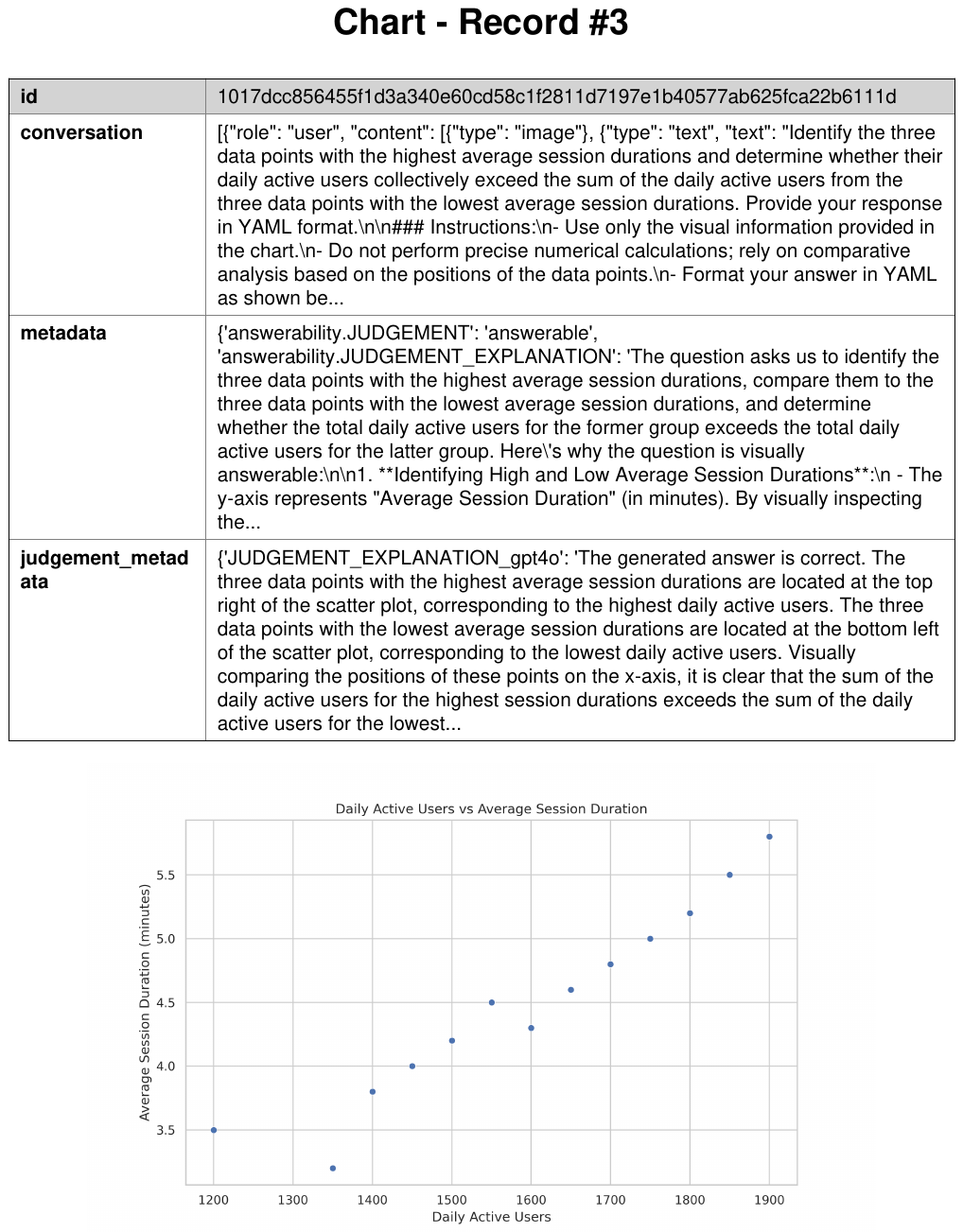}
\end{figure}

\clearpage

\end{document}